\documentclass[conference]{IEEEtran}
\IEEEoverridecommandlockouts

\usepackage{cite}
\usepackage{amsmath,amssymb,amsfonts}
\usepackage{graphicx}
\usepackage{textcomp}
\usepackage[ruled,vlined]{algorithm2e}
\usepackage{subcaption}
\usepackage[normalem]{ulem}
\usepackage{multirow}
\usepackage{soul}
\newcommand{\ie}{\textit{i.e.}}

\usepackage{comment}
\usepackage{booktabs}
\newcommand{\CP}[1]{\ignorespaces}

\def\BibTeX{{\rm B\kern-.05em{\sc i\kern-.025em b}\kern-.08em
    T\kern-.1667em\lower.7ex\hbox{E}\kern-.125emX}}

\begin{document}

\title{Reducing the False Positive Rate Using Bayesian Inference in Autonomous Driving Perception\\
}

\author{\IEEEauthorblockN{~Gledson Melotti$^{1}$}
	\and
	\IEEEauthorblockN{~Johann~J.~S.~Bastos$^1$}
	\and
	\IEEEauthorblockN{~Bruno~L.~S.~da~Silva$13$}
	\and
	\IEEEauthorblockN{~Tiago~Zanotelli$1$}
	\and
	\IEEEauthorblockN{~Cristiano~Premebida$^2$}
	
	\thanks{$^1$ Gledson~Melotti, Johann~J.~S.~Bastos, Bruno~L.~S.~da~Silva and Tiago~Zanotelli are with Federal Institute of Espirito Santo, Brazil. E-mail: \footnotesize \{gledson,bruno.legora,tiagoz\}@ifes.edu.br and jjakobschmitz@gmail.com}%
	
	\thanks{$^{2}$ C.Premebida is with the Institute of Systems and Robotics (ISR-UC), and the Dep. of Electrical and Computer Engineering at University of Coimbra-Portugal. E-mail: \footnotesize cpremebida@isr.uc.pt}%
}

\maketitle

\begin{abstract}
Object recognition is a crucial step in perception systems for autonomous and intelligent vehicles, as evidenced by the numerous research works in the topic. In this paper, object recognition is explored by using multisensory and multimodality approaches, with the intention of reducing the false positive rate (FPR). The reduction of the FPR becomes increasingly important in perception systems since the misclassification of an object can, depending on the circumstances, potentially cause accidents. In particular, this work presents a strategy through Bayesian inference to reduce the FPR considering the likelihood function as a cumulative distribution function from Gaussian kernel density estimations, and the prior probabilities as cumulative functions of normalized histograms. The validation of the proposed methodology is performed on the KITTI dataset using deep networks (DenseNet, NasNet, and EfficientNet), and recent 3D point cloud networks (PointNet, and PointNet++), by considering three object-categories (cars, cyclists, pedestrians) and the RGB and LiDAR sensor modalities.
\end{abstract}


\section{Introduction}
Research on sensory perception has achieved very satisfactory results in terms of object recognition, contributing significantly to the progress of autonomous and intelligent vehicles (AV/IV) and robotics, due to technological advances such as hardware, sensors and statistical learning techniques~\cite{SHI23,yanli,HE2023}. Perception systems for AV/IV can be understood as a process that interprets the data provided by the sensors in order to understand the surrounding environment, thus contributing to safer decision-making. An important item in perception systems is the object classification part, which is currently dominated by deep network (DN) architectures~\cite{Singh,Liao,He,Janai}.
\begin{figure}[!t]
	\begin{center}
		\includegraphics[width=0.479\textwidth]{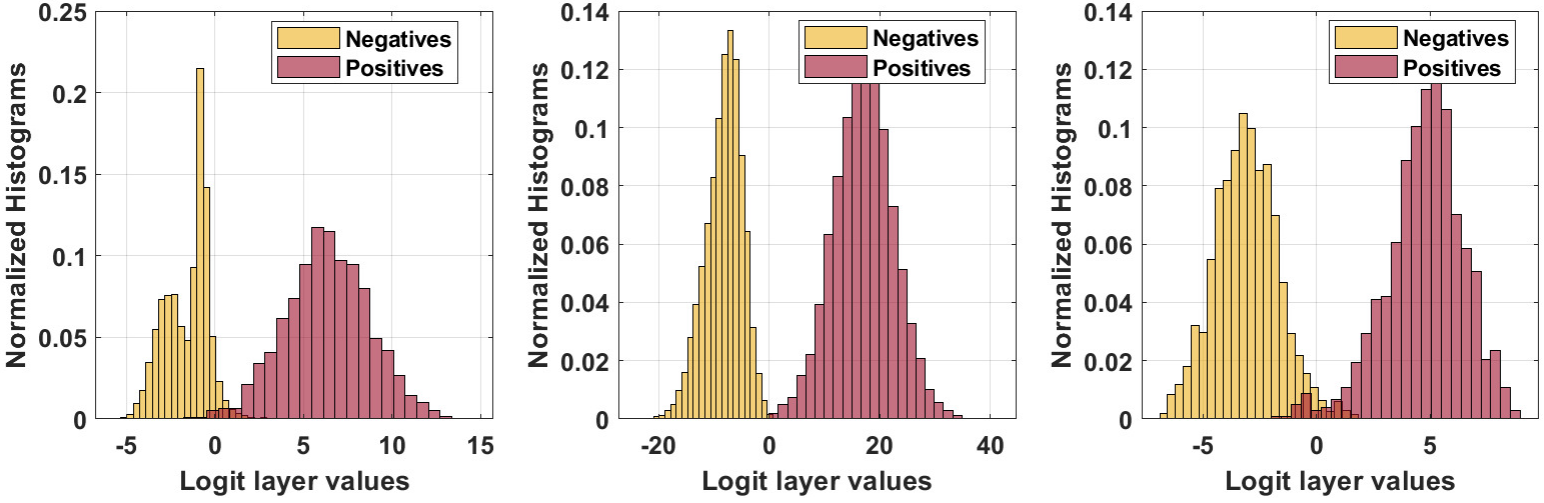}\\
		\includegraphics[width=0.479\textwidth]{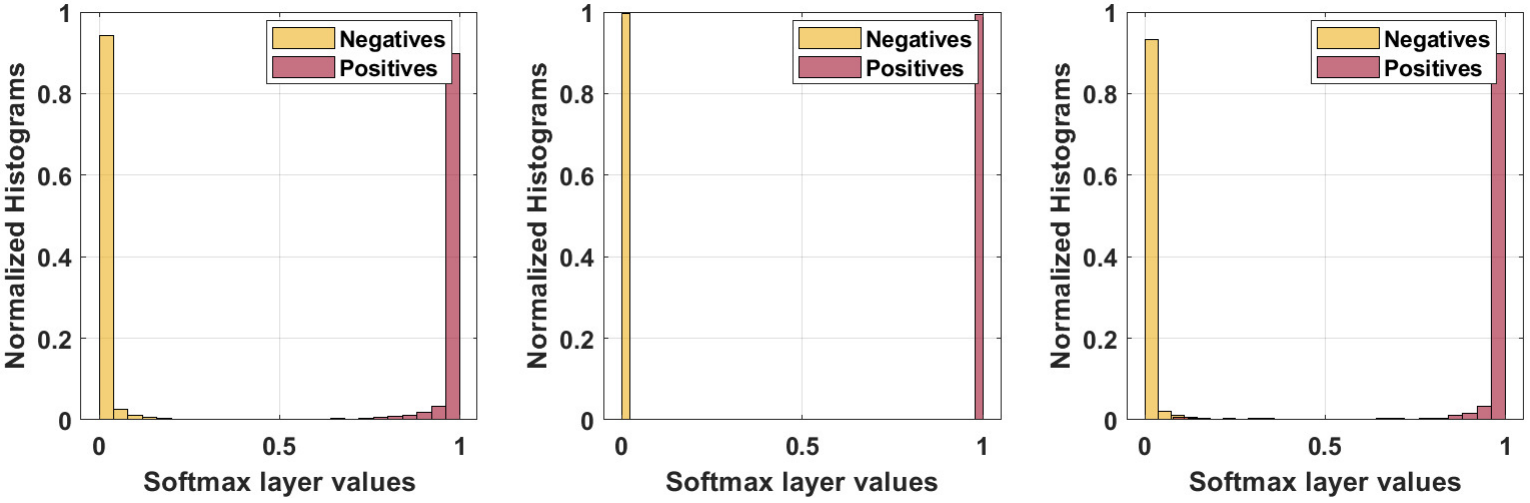}
		\caption{Representations of distributions in normalized histograms (NH) of logit values in the 1st row, and the 2nd row the softmax values.} 
		\label{HG_RGB_Johann}
	\end{center}
\end{figure}
\noindent\raggedbottom

Frequently, the DNs output the predictions as normalized scores, between $0$ and $1$, by using the Softmax or the Sigmoid functions~\cite{Guo,Su,Gasparovic}. However, DNs tend to be overconfident and do not always correctly classify the objects. The misclassified objects, false positives (FPs) or missing (FNs), hinder proper decision-makings by the perception systems. Thus, the reduction of the FPs in classification systems would provide safer actions for decision-making, especially in autonomous robots and intelligent vehicles applications~\cite{hogan,HE2023,MAHAUR2023}.

An alternative to reduce the FPR can be through probabilistic explainable approach by observing the logit layer values (score values before the prediction layer). Figure~\ref{HG_RGB_Johann} shows the distribution of logit values from an already trained network, in the first row, while the second row shows the distribution of the same scores after the softmax. It is possible to see that the logit values are smoother than the softmax values \ie, the values from softmax function are extreme (many values close to zero and many values close to one)~\cite{Guo,Pereyra,melotti1,melotti2}.

Softmax function ($SM$) is generally used as prediction function to classify a given input (decision-making)~\cite{TOVIAS2023,Guo}. Such function can be taken in the probabilistic context by means of the Bayes' theorem, considering probabilistic generative models (Naive Bayes, Bayesian networks and Hidden Markov Models) where the class-conditional densities and the prior are modeled (or known), and then the posterior probabilities can be estimated through the Bayes' theorem. In other words, first we have to determine the class-conditional density (likelihood function) for each class individually and then the class prior probability. Equivalently, the joint distribution can be directly modeled and later normalized to obtain the posterior probability. In fact, the classification result is given through two stages, the first being inference (distribution modeling) and the second being decision-making (classification)~\cite{Bishop}.

Alternatively, many traditional approaches to classification problems are of the type called discriminative models (logistic regression and support vector machine) or discriminant functions (traditional neural networks and k-nearest neighbors). The first tries to model a posterior probability directly using a parametric model in the inference stage, and consequently optimizing the parameters on the training set. Given the posterior model, for each new entry, it assigns a top-class label. The second case \ie, discriminant function, the approach defines a function which uses the training data to map each entry directly to a certain class (input-output mapping), and according to~\cite{Bishop} the ``probabilities play no role'' \ie, it is not possible to access posterior probabilities. In this case the inference and decision stages are into a single learning algorithm~\cite{Bishop}. 

The result achieved by a machine learning algorithm, such as a classifier considering predictions after a $SM$, should be carefully analyzed, so that the prediction result may not be considered as a proper probabilistic value \textit{per se}. To obtain adequate probabilistic results, the structure of the learning algorithms must encompasses probabilistic formulations.

In this context, this paper explores the well-known Bayes' theorem as a probabilistic interpretation of the predicted values from the logit values, through Maximum Likelihood (\textit{ML}) and Maximum a-Posteriori (\textit{MAP}) formulations. Additionally, we aim to reduce the false positive rate ($FPR$) without degrading the results already achieved by the neural networks. In fact, the ML and MAP formulations replace the predicted values of the neural network trained with the $SM$ prediction function, without the need to retrain the network \ie, the likelihood functions and prior probabilities of each class were obtained with the logit values (before $SM$ prediction layer). The likelihood function is then defined by as a cumulative distribution function (CDF) from the Gaussian kernel density estimation, and the prior probability as a cumulative function from of normalized histogram (NH), both with the logit values of trained networks.

In summary, the contributions are:

\begin{itemize}
	\item [$\bullet$] An investigation of the parametric and nonparametric modeling to represent the likelihood function and the prior probabilities, considering CDFs;
	\item [$\bullet$] Reducing the FPR through Maximum likelihood and Maximum a-Posteriori formulations for object classification;
	\item [$\bullet$] A study with five distinct neural network architectures to validate the proposed approach, taking into account datasets from different modalities and sensors (RGB images, and $3D$-LiDAR point clouds), contributing to advances in multisensory and multimodality perception.
\end{itemize}

\section{Related work}
There are many recent works that address False Positive Reduction techniques in different contexts, such as disease detection, security breach detection and vehicle detection. For the first context, the authors in~\cite{3darcnn} proposed a novel asymmetric residual network that uses 3D features and spatial information to improve classification and reduce false positives in lung nodule detection. Their network showed promising results in reducing false positive in clinical applications. In~\cite{reducingFPR}, the authors proposed a post-processing method to estimate the confidence score of the predictions from a single channel CNN architecture. While using the confidence score of several layers of a CNN, their approach could reduce up to 18\% of the false positive detection in one of their tests. The authors of~\cite{mhsnet} proposed a post-processing method to reduce false positives in lung cancer detection. Their method is lightweight and does not bring any constraints in the ``front network'', while reducing 6.4\% of the false discovery rate in their tests. In~\cite{slicefusion}, the authors proposed a novel slice-fusion method with a Mask R-CNN detection model to reduce false positives in liver tumor detection and segmentation.

Regarding the security breach context, the authors of~\cite{intrusion} proposed a framework for addressing zero-day attacks in software (attacks that occur before the developer can take action on it), combining features selection methods and fine-tuning of their datasets. In~\cite{GNJ}, the authors proposed a technique to reduce false positives in hardware Trojan (HT) detection. Their method combines signal justification and unsupervised K-means, and is a general technique that can be applied to suspicious signals in detecting HT. Their experiments were done on various combinations of full and partial-scans of circuits, and obtained a false positive ratio of 3.89\% and 3.31\% for full and partial scans of circuits. In~\cite{hybridintrusion}, an improved stacking ensemble algorithm was proposed to enhance the true positive rate of a intrusion detection system (IDS). Their Hybrid IDS was tested and their results showed that the method was superior than the compared techniques, in terms of True and False positives. In the software development context, the authors of~\cite{bugdetection} proposed a Transformer-based learning approach to identify false positive bug warnings found by static analysis tools, which usually return a large number of false positives that developers must verify manually. Their approach improved the precision of a tool by 17.5\% and 5.5\%, when considering \textit{null dereferences} and \textit{resource leaks} warnings, respectively.

Lastly, in the vehicle detection context, the authors of~\cite{lateassymetric} proposed an asymmetric late fusion approach to combine camera and LiDAR outputs from different networks. Their objective was to eliminate false positives in these object detectors. According to their results, their objective was attained and the method achieved up to 9.87\% better class-wise performance than the LiDAR-only detector. In~\cite{maff}, two end-to-end trainable feature fusion techniques were proposed to combine RGB and point-cloud features. Their experiments showed that their methods can improve significantly the filtering of false positive from data. Their approaches can be applied to improve the false positive ratio in many different architectures. The paper in~\cite{hogan} addresses the influence of images from a fisheye camera \ie, such cameras can include undesirable parts of the vehicle's ego body in the perception system of autonomous vehicles, as well as the reflections of objects on car bodies, where both can produce false positives, and reduce the efficiency of object detection systems. Thus, the authors proposed a neural network architecture to identify and extract the vehicle's ego-body. Put another way, eliminating the possibility of pedestrians or other objects being wrongly detected in the car's ego-body reflection. In this way, the authors showed a reduction in false positives by eliminating the vehicle’s ego body with the reflected objects.

\section{Proposed Method}
\label{pro_met}

\subsection{Probabilistic Inference}
\label{Probabilistic_Inference}

This section presents the formulations to reduce $FPR$, through Maximum Likelihood (\textit{ML}) and Maximum a-Posteriori (\textit{MAP}) functions, based on the Bayes' rule \eqref{bayes0}, including nonparametric and parametric modeling to define the posterior probability, likelihood function, and prior probability as well. Expressing the posterior by
\begin{align}
	P(\mathbf{C}|\mathbf{Sc})=\cfrac{P(\mathbf{Sc}|\mathbf{C})P(\mathbf{C})}{P(\mathbf{Sc})},
	\label{bayes0}
\end{align}
where $\mathbf{C}$ is the random variable (RV) associated to the object categories, $\mathbf{Sc}$\footnote{Generally, neural network score values are obtained using a prediction function that normalizes logit values between zero and one, such as the softmax prediction function.} are the classified object scores (predicted values), $P(\mathbf{Sc}|\mathbf{C})$ is the likelihood, $P(\mathbf{C})$ is the prior probability, and $P(\mathbf{Sc})\neq 0$ is the model evidence, considering the pior and likelihood are known. From the Law of Total Probability~\cite{Bishop}, (\ref{bayes0}) can be rewritten using the \textit{per-class} expression,
\begin{align}
	P(c_i|\mathbf{Sc}) = \cfrac{P(\mathbf{Sc}|c_i)P(c_i)}{\sum\limits_{i=1}^{nc}P(\mathbf{Sc}|c_i)P(c_i)},
	\label{bayes1}
\end{align}
where $P(\mathbf{Sc}|c_i)$ is the likelihood of an object for the class ($c_i$). Given (\ref{bayes1}), an inference can be made on the test set about the ``unknown'' RV $\mathbf{C}$ from the dependence with $\mathbf{Sc}$ \ie, the value of the posterior distribution of $\mathbf{C}$ is determined 
from $\mathbf{Sc}$~\cite{melotti1,melotti2}.

\subsection{ML and MAP Functions}
\label{ml_map_function}

We argue that the values from the logit layer are more suitable for representing a probability density function, when compared with the values of the $SM$ function, as illustrated by the distributions in Fig. \ref{HG_RGB_Johann}. Thus, from (\ref{bayes1}) we can define the Maximum Likelihood and the Maximum a-Posteriori function, as (\ref{bayes2}) and (\ref{bayes3}) respectively~\cite{melotti1,melotti2}:
\begin{align}
	ML := arg \max_{i} \cfrac{(P(\mathbf{\textbf{Sc}}|c_i)+\lambda)}{\sum\limits_{i=1}^{nc}(P(\mathbf{\textbf{Sc}}|c_i)+\lambda)},
	\label{bayes2}
\end{align}

\begin{align}
	MAP := arg \max_{i} \cfrac{(P(\mathbf{\textbf{Sc}}|c_i)P(c_i)+\lambda)}{\sum\limits_{i=1}^{nc}(P(\mathbf{\textbf{Sc}}|c_i)P(c_i)+\lambda)}
	\label{bayes3},
\end{align}
where $\lambda$ represents the additive smoothing parameter, used here to avoid the zero probability problem~\cite{Valcarce}.

\subsection{Kernel Density Estimation and Normalized Histogram}
\label{K_D_E_N_H}

\begin{figure}[!tb]
	\begin{center}
		\includegraphics[width=0.23\textwidth]{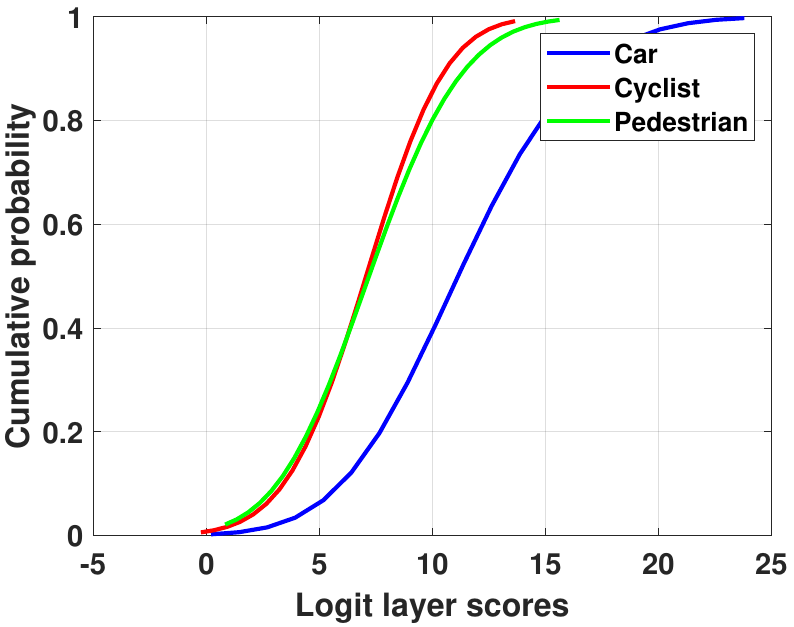}
		\includegraphics[width=0.227\textwidth]{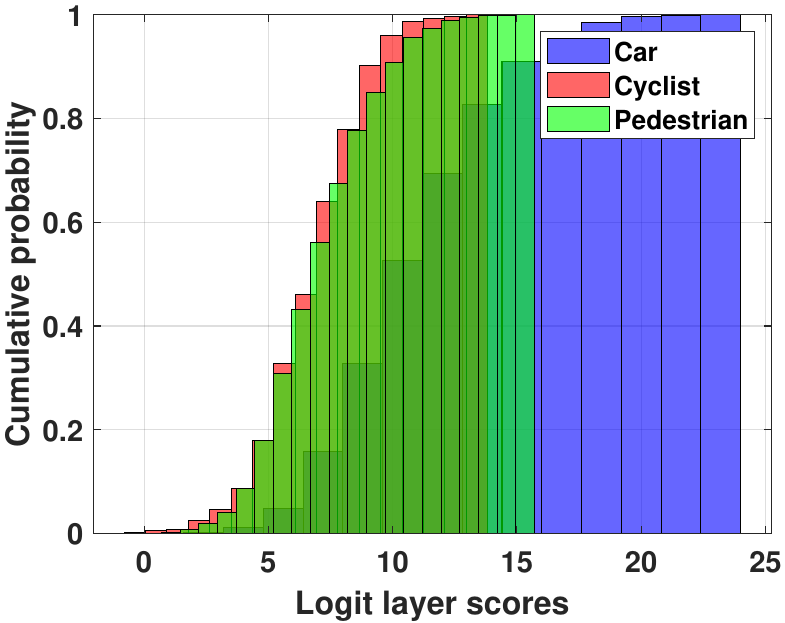}
		\caption{CDF representations obtained from Gaussian functions (first column) and normalized histograms (second column).}
		\label{pdf_cdf}
	\end{center}
\end{figure}

We propose to model the \textit{ML} and \textit{MAP} functions, which will perform the inference, by taking the CDF, where the density is extracted by using the logit layer values (\ie, before softmax values) from the training set data, as illustrated in Fig. \ref{pdf_cdf}.
\noindent
\raggedbottom

The curves on the left hand-side of Fig. 2 represent CDFs modelled by Gaussians (likelihood funtions) \ie, they are modeled using parametric estimates by means of a Kernel Density Estimation (KDE) given in (\ref{kde_1}) by
\begin{align}
	\widehat{f}_{ker}(d) = \frac{1}{nh}\sum_{i=1}^{n}\frac{1}{\sqrt{2\pi}}e^{-{\left(\frac{1}{2h^2}\right)}{\left(d-Sc_i\right)}^2},
	\label{kde_1}
\end{align}
where $h$ is a smoothing parameter\footnote{This smoothing parameter is not related to the smoothing parameter of Bayesian inference functions (\textit{ML} and \textit{MAP}).} called window width or bandwidth (bw)\footnote{Small values lead to rough curves, larger values lead to smoother curves.}, $n$ is the number of observations, $d$ is a value set\footnote{The $d$ values are not related to the values of scores or logits.} (domain) that evaluates the function $\widehat{f}_{ker}(d)$, $Sc_i$ are the predicted values (scores) of each object classified to a certain class. Density estimation is obtained by computing the average of several probability density functions from \eqref{kde_1}, considering the set of values $d$, and consequently obtain a CDF~\cite{ScottDavidW,Martinez,melotti1}.

The idea of applying Gaussian functions is to obtain a smoother distribution, as shown in Fig. \ref{HG_RGB_Johann} (see the 1$^{st}$ row). In other words, the distribution from the logit layer is more suitable for modeling a probability density function. Furthermore, the Gaussian distribution has a maximum entropy \ie, a distribution with more information and less confident information around the mean (distribution with high variance)~\cite{Bishop,melotti2}.

Prior probabilities are represented by CDFs obtained from normalized histograms (NH), as illustrated on the right of Fig \ref{pdf_cdf}. According to ~\cite{Martinez} ``Histograms are a good way to $i)$ summarize a data set to understand general characteristics of the distribution such as shape, spread, or location; $ii)$ suggest possible probabilistic models, $iii)$ or  determine unusual behavior''. In other words, here the NH is used to model proper distributions. The histogram is constructed from the number of bins (intervals) \ie, the number of bars, which must not overlap with each other and the bins should have the same width~\cite{melotti1}.

\begin{figure}[!t]
	\begin{center}
		\includegraphics[width=0.48\textwidth]{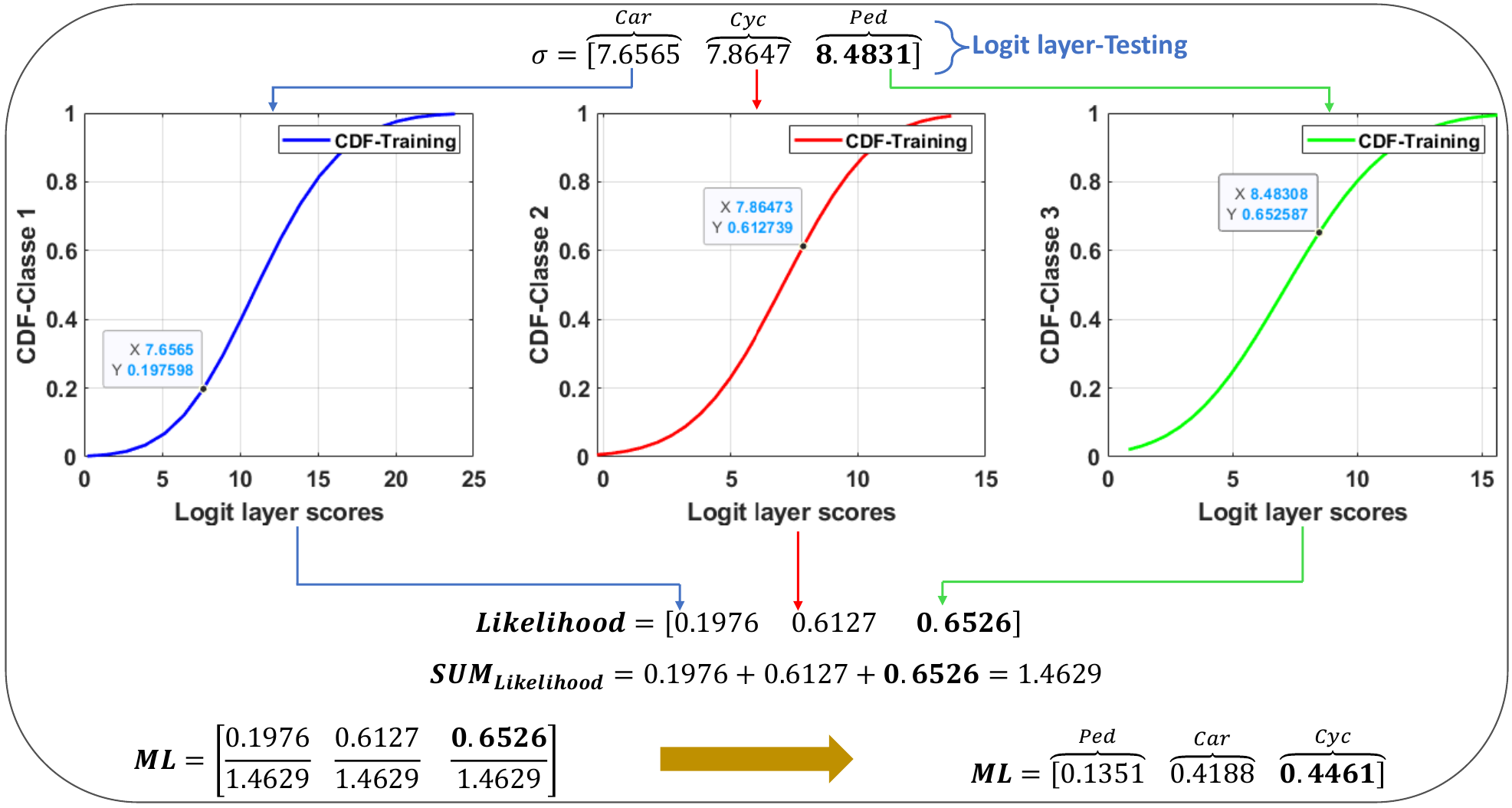}
		\caption{The likelihood function ($P(\mathbf{\textbf{Sc}}|c_i)$) is calculated per class for each classified object \ie, the \textit{ML}.}
		\label{ML_CDFs_Por_Classes_Final}
	\end{center}
\end{figure}

\begin{figure}[!t]
	\begin{center}
		\includegraphics[width=0.48\textwidth]{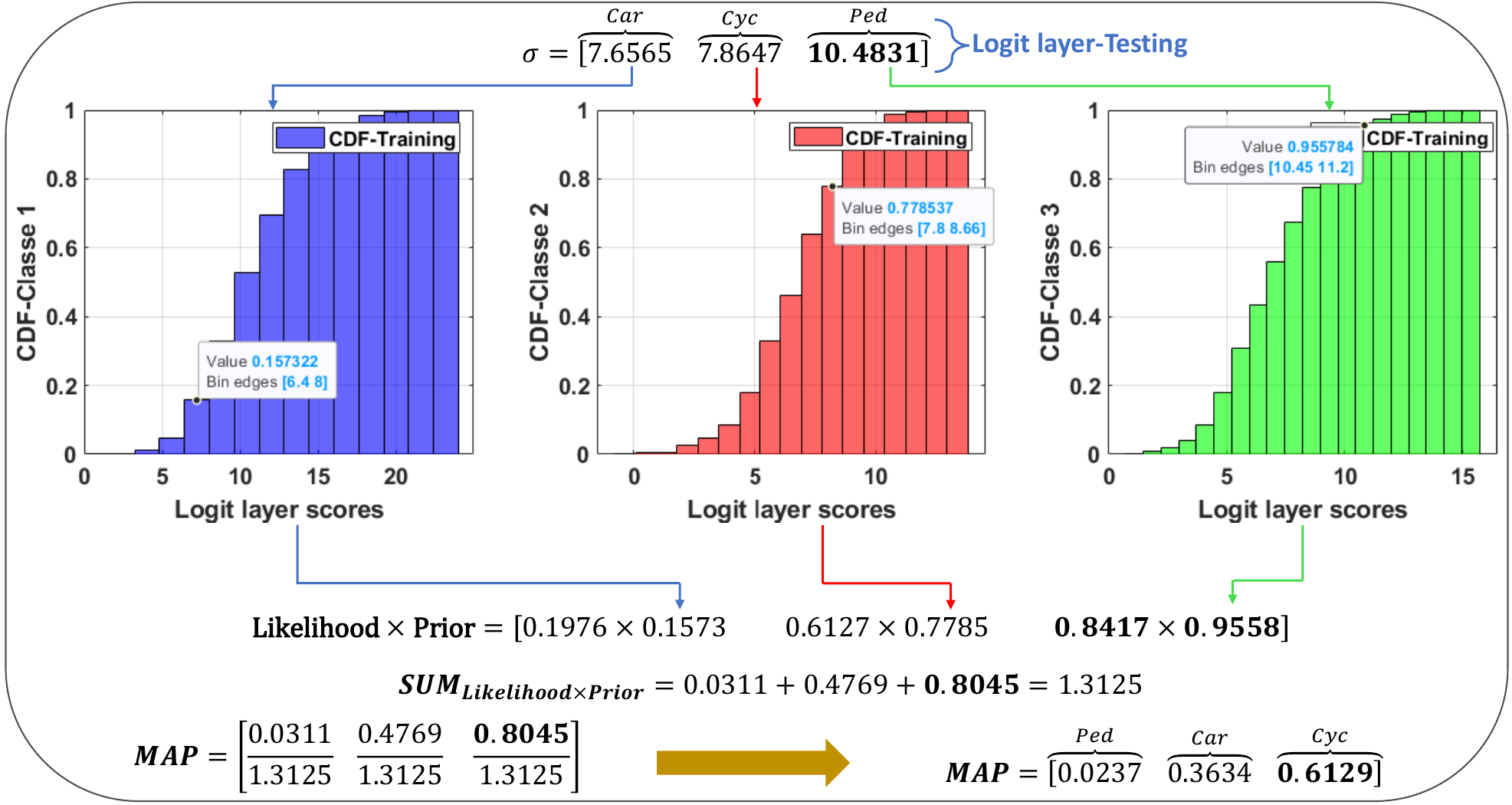}
		\caption{The prior probability ($P(c_i)$) is computed per class for each classified object, as well as the \textit{MAP}.}
		\label{MAP_CDFs_Por_Classes_Final}
	\end{center}
\end{figure}

In an implementation perspective, the formulation to get $P(\mathbf{\textbf{Sc}}|c_i)$ for the \textit{ML} function can be computed as illustrated in Fig. \ref{ML_CDFs_Por_Classes_Final}, while the \textit{MAP} function (posterior probability) follows as illustrated in Fig. \ref{MAP_CDFs_Por_Classes_Final}. Therefore, the \textit{ML} and \textit{MAP} functions replace the softmax function only on the test data, using the logit layer values, while the CDFs were obtained with the logit data from the training dataset. Notice that, although the Bayesian formulation takes distributions into account, \textit{ML} and \textit{MAP} compute a \CP{single}point estimate rather than a distribution.
\noindent
\raggedbottom

\noindent\raggedbottom

The use of different models to represent the distributions aims to capture different information from the training data. The choice of obtaining a CDF from a Gaussian distribution to represent the likelihood function and a CDF from NH to represent the prior probability were defined based on preliminary experiments. The reverse could be valid \ie, a Gaussian distribution for the likelihood and NH for the prior probability.

\subsection{Setting the KDE and NH Parameters}
KDE's formulation involves determining $\lambda$ parameters for \textit{ML} and \textit{MAP} functions, as well as $h$ (smoothing parameter) for each class, according to (\ref{bayes2}), (\ref{bayes3}), and (\ref{kde_1}). Differently, the NHs are constructed using the number of bins ($nbins$) for each class. Thus, the determination of such parameters
were obtained through a genetic algorithm\footnote{In this work, the Matlab genetic algorithm toolbox was used.}, considering in the cost function the F-score (F1) and $FPR$ metrics, as defined in (\ref{ga}),
\begin{align}
	F_{cost} = min [(1-F1) + FPR].
	\label{ga}
\end{align}

The parameters $\lambda$ and $h$ were determined by considering a subset of $\mathbb{R}$ as search space, while $nbins$ were determined by having only integers in the search space. The optimization process of the genetic algorithm was carried out with the training data and validated on the validation data, for the determination of the KDE and NH parameters \ie, the parameters that provide the lowest value for FPR and the highest value for the F-score. The internal parameters of the genetic algorithm were crossover fraction equal to $0.8$, maximum generation equal to $100$ times the number of variables, population size equal to $200$ and the mutation was determined by applying random number chosen from a Gaussian distribution, to each entry of the parent vector.

Note that this paper aims to reduce the rate of false positives without degrading the classification results, in other words, without degrading the F-score metric: this is the reason of using the F-score in the cost function of the genetic algorithm.

\subsection{Dataset}
\begin{figure}[!t]
	\centering
	\includegraphics[width=0.98\linewidth]{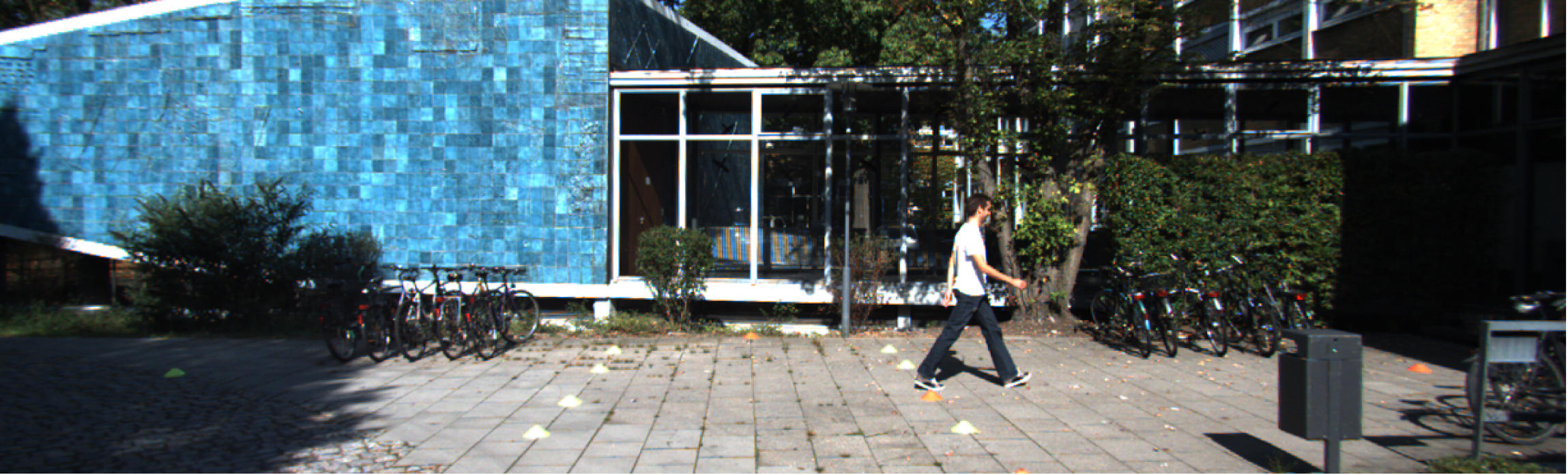}
	\includegraphics[width=0.98\linewidth]{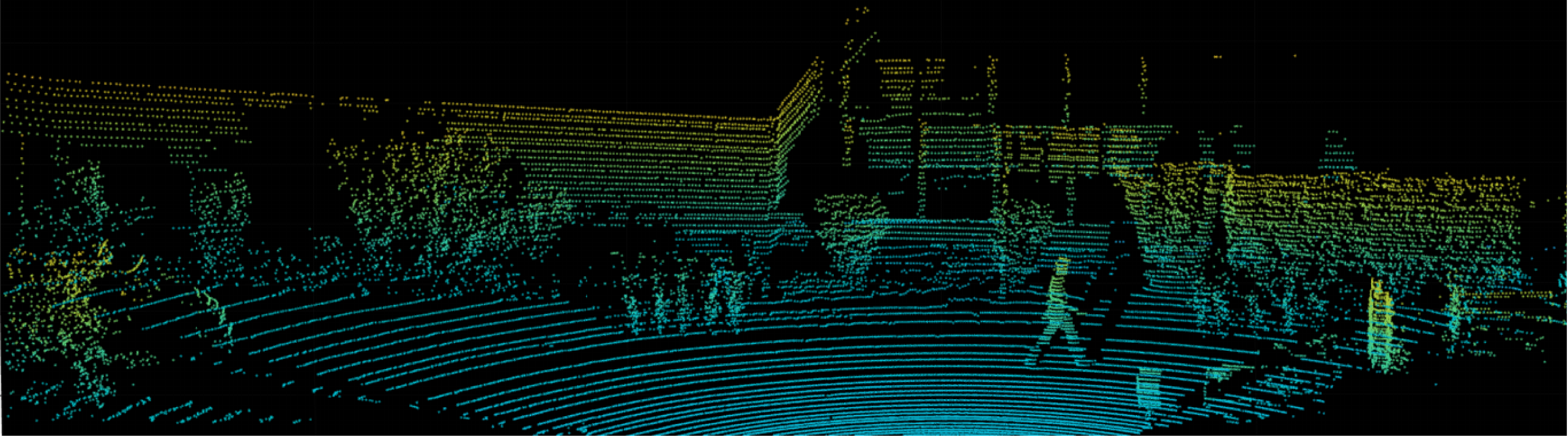}
	\includegraphics[width=0.98\linewidth]{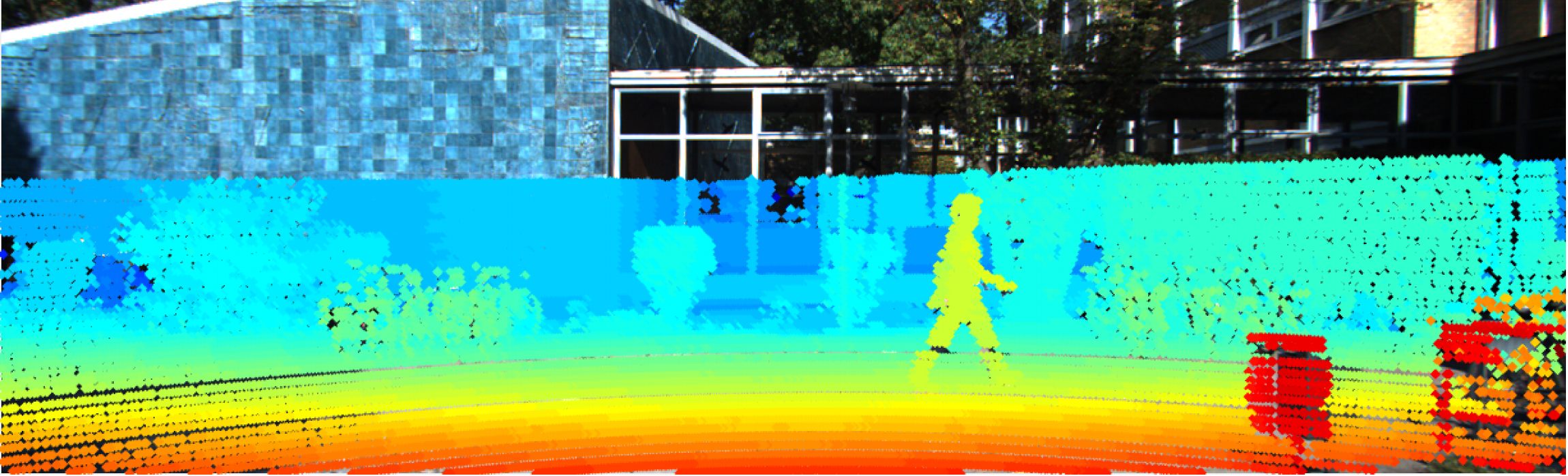}
	\includegraphics[width=0.98\linewidth]{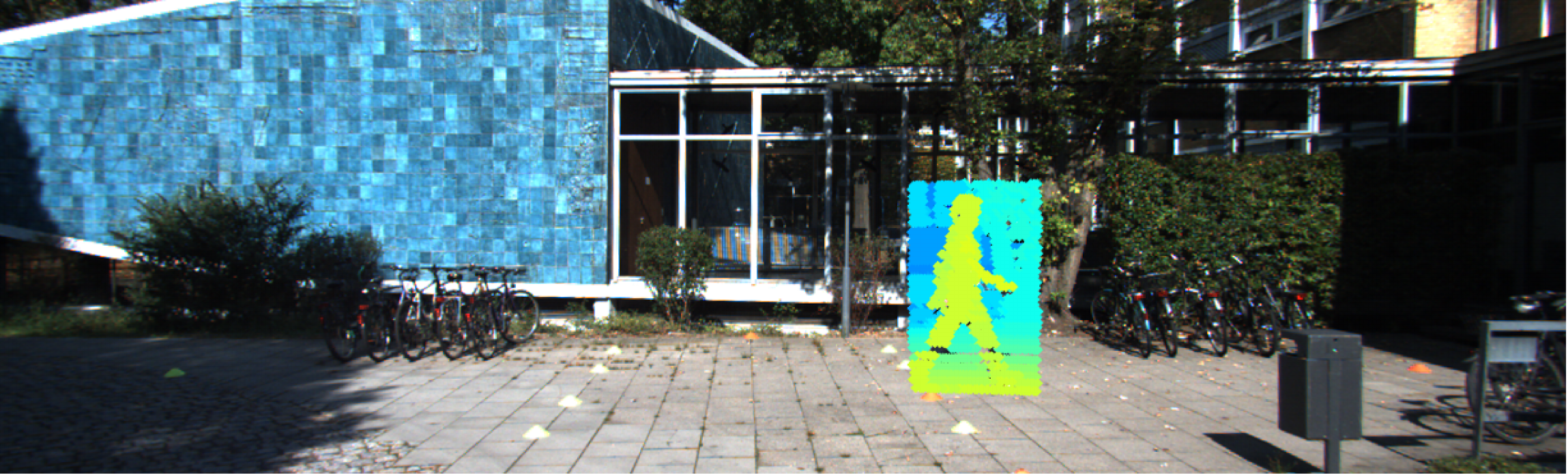}
	\caption{Example of an image obtained from a passive sensor - RGB camera (first row); $3D$ point clouds obtained from an active sensor - HDL-64E Velodyne $3D$ LiDAR (second row); projection of the $3D$ point clouds in the $2D$ image-plane (third row); and $3D$ point clouds projected just for the pedestrian object (fourth row). Images from the Object Detection Evaluation of the KITTI dataset.~\cite{geiger2012,geiger2013}}
	\label{projection}
\end{figure}

To validate the proposed methodology, this paper considers three neural networks that process RGB images (DenseNet~\cite{densenet}, NasNet~\cite{nasnet}, and EfficienteNet~\cite{efficientnet}) and two neural network that directly processes $3D$ point clouds (PointNet~\cite{pointnet}, and PointNet++~\cite{pointnet++}). The results were achieved using the KITTI Object Detection dataset, where the objects have been extracted (cropped) both from the RGB image frames and from the $3D$ point clouds frames, projecting the $3D$ points to the $2D$ image-plane~\cite{geiger2012,geiger2013}, as in Fig. \ref{projection} and Algorithm \ref{alg_3Dpc}. 

The point clouds projected on the image-plane are eliminated using the $2D$ bounding boxes of the $2D$ objects \ie, the projected points that are outside the $2D$ bounding boxes have their respective $3D$ points excluded from the $3D$ frame.

\begin{tiny}
	\begin{algorithm}[!t]
		\caption{Cropped $3D$ point cloud.} 
		\label{alg_3Dpc}
		\SetAlgoLined
		\KwIn{LiDAR sensor data and $2D$ bounding boxes.}
		\KwOut{Cropped $3D$ point clouds.}
		\textbf{Getting the $\mathbf{3D}$ point clouds}\\
		$pc \gets OpenLiDAR(data)$;\\
		$indices \gets pc(:,1)<5$; /* Points that do not belong to the $2D$ image-plan are removed (the value is an approximation) */ \\
		$pc(indices,:) \gets [\; \;]$;\\
		\textbf{Project PC for image-plane}\\
		$pc_{proj} \gets P_{rect}R_{rect}T_{LiDAR}^{Cam}PC$;\\
		$pc_{proj}(:,1) \gets pc_{proj}(:,1)/pc_{proj}(:,3)$;\\
		$pc_{proj}(:,2) \gets pc_{proj}(:,2)/pc_{proj}(:,3)$;\\
		\textbf{Defining the points inside the bounding box}\\
		$Boxes = [x_{min}\quad y_{min}\quad x_{max}\quad y_{max}]$;\\
		$indices \gets [\; \;]$;\\
		\For{$i \gets 1:Size(pc_{proj})$}{
			\uIf{$(pc_{proj}(i,1) >= Boxes(1) \; {\bf and} \; pc_{proj}(i,1) <= Boxes(3)+1) \; {\bf and} \; (pc_{proj}(i,2) >= Boxes(2) \; {\bf and} \; pc_{proj}(i,2) <= Boxes(4) +1)$}{
				$indices \gets [indices;  \; i]$
		}}
		$PC=pc(indices,:)$;\\
	\end{algorithm}\noindent\raggedbottom
\end{tiny}

\begin{figure}[!t]
	\centering
	\includegraphics[width=1.0\linewidth]{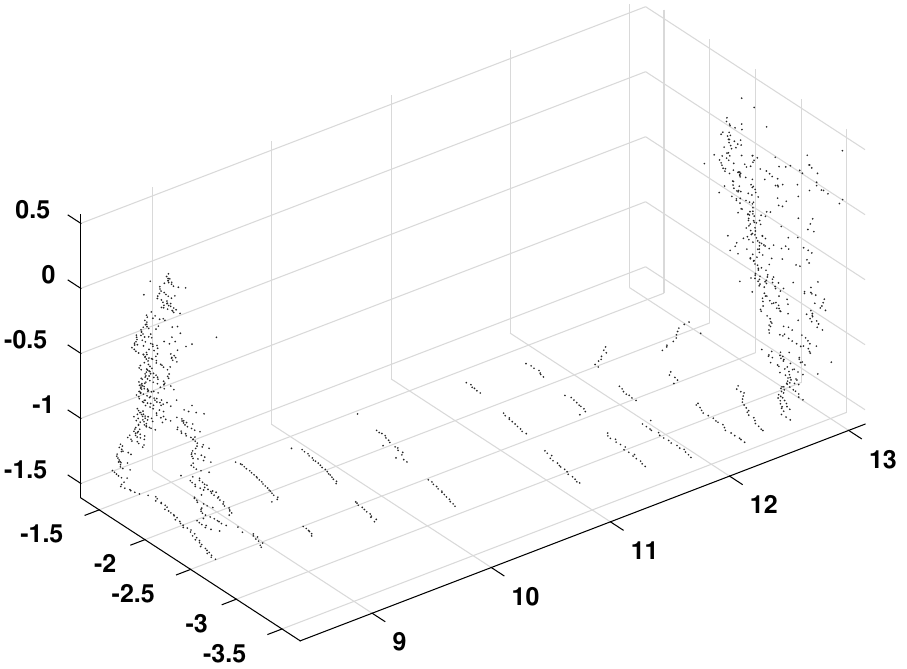}
	\caption{Example of cropped $3D$ object, presenting background points.}
	\label{Pedestrian_Point_Cloud_paper}
\end{figure}

It cannot be overlooked that the LiDAR sensor's operating principle is the reflection of light beams \ie, objects are generated by light reflections. Like this, there are $3D$ points that do not belong to the cropped $3D$ objects therefore, such points are defined as backgrounds or foregrounds points. The Fig. \ref{Pedestrian_Point_Cloud_paper} illustrates an pedestrian object with background points, from the $3D$ points projected according to the last row of Fig. \ref{projection}.

The backgrounds or foregrounds points were removed via a clustering technique based on the distance between points, to define the points belonging to the objects, as shown in Alg. \ref{alg_Cluster}.
\begin{tiny}
	\begin{algorithm}[!t]
		\caption{Cluster.} 
		\label{alg_Cluster}
		\SetAlgoLined
		\KwIn{Cropped $3D$ point cloud and distance between points.}
		\KwOut{$3D$ point clouds with a cluster.}
		\textbf{Compute the Euclidean distance}\\
		$reference \gets 0$;\\
		$Dist_{pc} \gets EucDistance(PC_{WithoutCluster},reference)$;\\
		$indice \gets [1:1:Size(PC_{WithoutCluster})]$;\\
		$Dist \gets [Dist_{pc} \quad indice]$;\\
		$Dist \gets SortRows(Dist,1)$;  \\
		\textbf{Compute the cluster}\\
		$distance \gets 0.25$;\\
		$id_{cluster} \gets Zeros([Size(PC_{WithoutCluster}), \; 1])$\\
		$id_{master}(1) \gets 1$;\\
		\For{$i \gets 2:Size(PC_{WithoutCluster})$}{
			\uIf{$Dist(i,1)-Dist(i-1,1) <= distance$}
			{
				$id_{cluster} \gets id_{master}$;}
			{$id_{master} \gets id_{master}+1$;\\
				$id_{cluster} \gets id_{master}$;}
		}
		\textbf{Check the cluster and compute the histogram count}\\
		$Cluster \gets Unique(id_{cluster})$;\\
		$HC \gets HistogramCount(id_{cluster},Cluster)$;\\
		$confidence \gets 1$ /* Confidence level */\\
		$cl \gets Size(HC)$ /* Number of clusters in the sample */\\
		\lFor{$i \gets 1:Size(cl)$}{
			$HC(i) \gets HC(i)*\left(confidence-\frac{(i-1)}{cl}\right)$
		}
		$[ClusterCount \quad Position] \gets Max(HC)$;\\
		$ct \gets 1$;\\
		\For{$i \gets 1:Size(id_{cluster})$}{
			\uIf{$id_{cluster}(i)==Cluster(Position)$}
			{$PC_{Cluster}(ct,:) \gets PC_{WithoutCluster}(Dist(i,2),:)$;\\
				$ct \gets ct+1$;}
		}
	\end{algorithm}\noindent\raggedbottom
\end{tiny}

\begin{figure}[!t]
	\begin{center}
		\includegraphics[width=0.23\textwidth]{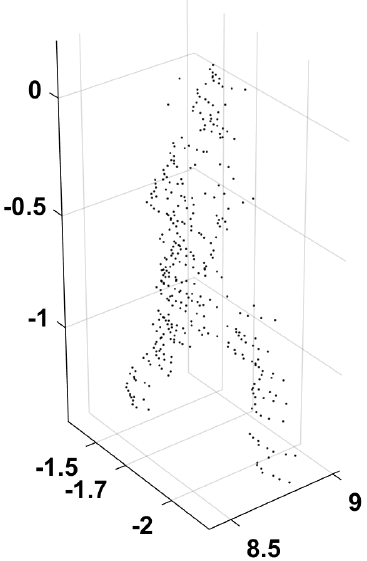}
		\hspace{0.2cm}
		\includegraphics[width=0.227\textwidth]{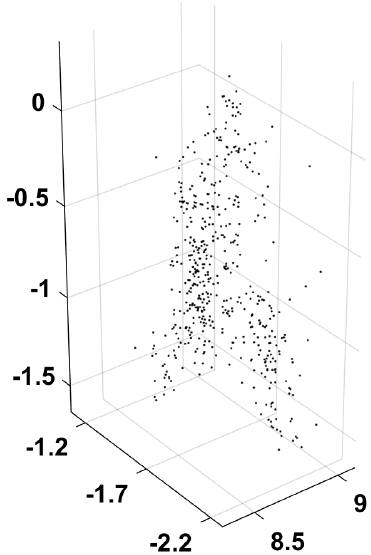}
		\caption{The pedestrian on the left contains the number of original points of the frame (364 points), on the right the same pedetrian with 512 points (upsample). The axis are shown in meters.\CP{Point clouds are defined by distances given in meters.}}
		\label{pedestrian_crop}
	\end{center}
\end{figure}

The $3D$ point clouds objects contain different amount of points because of the nature of the $3D$ LiDAR sensor. Thus, some objects had the amount of points reduced to 512 (random downsample) or increased to 512 points (considering the k nearest neighbors to sample the $3D$ points), as shown in Fig. \ref{pedestrian_crop}. Table \ref{dataset} shows the number of objects for images and point clouds modalities. The number of objects between $2D$ images and $3D$ point clouds are different, as the LiDAR sensor has distance limitations \ie, some objects are not captured by \CP{such a}the sensor.
\noindent\raggedbottom

\begin{table}[!t]
	\begin{small}
		\begin{center}
			\caption{KITTI dataset for classification: number of objects per class and and respective subsets.}
			\label{dataset}
			\begin{tabular}{cccc}
				\toprule
				\multicolumn{4}{c}{ \textbf{RGB Images - 7481 Frames} } \\
				& \textbf{Car} & \textbf{Cyclist} & \textbf{Pedestrian} \\ \hline
				\textbf{Training}    & $18103$ & $1025$  & $2827$ \\
				\textbf{Validation}  & $2010$  & $114$   & $314$ \\
				\textbf{Testing}     & $8620$  & $488$   & $1346$ \\
				\hline
				\multicolumn{4}{c}{ \textbf{$3D$ Point Clouds - 7481 Frames} } \\
				& \textbf{Car} & \textbf{Cyclist} & \textbf{Pedestrian} \\ \hline
				\textbf{Training}    & $15324$ & $923$  & $2688$ \\
				\textbf{Validation}  & $1717$  & $99$   & $303$ \\
				\textbf{Testing}     & $7332$  & $452$  & $1260$ \\
				\bottomrule
			\end{tabular}
		\end{center}
	\end{small}
\end{table}
\noindent\raggedbottom

\section{Experiments and Results}

Assessing the false positive rate (FPR) in real world applications is relevant in autonomous driving scenarios, since objects can be misclassified by 
neural networks with high score values. In this section we evaluated 
the proposed approaches by replacing the $SM$ prediction function by the \textit{ML} and \textit{MAP} ones only on the test set, considering the likelihood as CDFs from Gaussian functions, and prior probabilities are represented as CDFs obtained from NHs as described in Sect.\ref{pro_met}.
\begin{table*}[!t]
	\caption{Comparison between the classifications obtained by the \textit{SM}, \textit{ML} and \textit{MAP} functions in terms of average F-score and $FPR$ ($\%$).}
	\label{res_ml_map}
	\begin{small}
		\begin{center}
			\begin{tabular}{c|cc|cc|cc|cc|cc}
				\toprule
				\multirow{2}{*}{\begin{tabular}[c]{@{}c@{}}Prediction\\ Function\end{tabular}} & \multicolumn{2}{c|}{DenseNet} & \multicolumn{2}{c|}{NasNet} & \multicolumn{2}{c|}{EfficientNet} & \multicolumn{2}{c|}{PointNet} & \multicolumn{2}{c}{PointNet++} \\ \cline{2-11}
				& $FPR$ $\downarrow$    & F-score $\uparrow$ & $FPR$ $\downarrow$  & F-score $\uparrow$  & $FPR$ $\downarrow$     & F-score $\uparrow$ & $FPR$ $\downarrow$      & F-score $\uparrow$ & $FPR$ $\downarrow$      & F-score $\uparrow$\\ \cline{1-11}
				$SM$        & $0.7202$ & $\mathbf{97.38}$ & $0.7202$ & $96.98$ & $0.4545$ & $\mathbf{98.44}$ & $5.36$ & $\mathbf{80.39}$ & $2.66$ & $\mathbf{92.51}$\\
				$ML_{KDE}$  & $\mathbf{0.6822}$ & $97.16$ & $\mathbf{0.7146}$ & $96.81$ & $\mathbf{0.3380}$ & $98.14$ & $\mathbf{4.45}$ & $80.00$ & $\mathbf{2.31}$ & $92.17$\\
				$MAP_{KDE}$ & $\mathbf{0.6748}$ & $97.05$ & $\mathbf{0.7026}$ & $\mathbf{97.08}$ & $\mathbf{0.3848}$ & $98.32$ & $\mathbf{4.49}$ & $80.10$ & $\mathbf{2.35}$ & $92.03$\\
				\bottomrule
			\end{tabular}
		\end{center}
	\end{small}
\end{table*}

Results achieved on the classification test set are shown in Table \ref{res_ml_map}, in terms of FPR and F-score 
measures. It can be seen that the FPR decreased after replacing the $SM$ function with  \textit{ML} and \textit{MAP}. Regarding the RGB image classifications, particularly with the EfficientNet network, the FPR decreased significantly, given that the reduction is $25.63\%$ for \textit{ML} and $15.34\%$ for \textit{MAP}. On the other hand, the values of the F-scores decreased very slightly in almost all networks (not compromising classification performance), with the exception of the NasNet network that increased the F-score using \textit{MAP} - which is very positive. Another significant FPR reduction occurred with the PointNet network, achieving a value of $16.98\%$ reduction when using \textit{ML}. The parameters ($nbins$, $\lambda$, and $bw$) determined by the genetic algorithm are presented in Tables \ref{par_ml}, \ref{par_map}, and \ref{par_bins_map}.

Finally, it is worth mentioning that this paper is not aiming to identify  which network is the best in terms of classification performance but, rather by proposing and evaluating a strategy to reduce the FPR \ie, the efficiency of the proposed methodology in networks that can potentially be employed as part of more reliable perception systems applied to robotics and autonomous vehicles. Furthermore, the research reported in this paper is not primary focused on developing a technique to eliminate background or foreground points nor a technique for sampling $3D$ point clouds to obtain a better classification result.

\begin{table}[!t]
	\caption{Parameter values for KDE \ie, the values of $h$ ($bw$) for each class, as well as the value of $\lambda$ in the \textit{ML} formulation.}
	\label{par_ml}
	\begin{small}
		\begin{center}
			\begin{tabular}{c|cccc}
				\toprule
				\multirow{2}{*}{\begin{tabular}[c]{@{}c@{}}Neural\\ Network\end{tabular}} & \multicolumn{4}{c}{Parameter} \\ \cline{2-5}
				& \multicolumn{1}{c|}{$h_{Car}$} & \multicolumn{1}{c|}{$h_{Cyc}$} & \multicolumn{1}{c|}{$h_{Ped}$} & $\lambda$ \\ \cline{1-5}
				DenseNet & \multicolumn{1}{c|}{$2.55$} & \multicolumn{1}{c|}{$9.60 \times 10^{-1}$} & \multicolumn{1}{c|}{$1.40$} & $2.54 \times 10^{-7}$ \\
				NasNet & \multicolumn{1}{c|}{$2.89$} & \multicolumn{1}{c|}{$4.95 \times 10^{-1}$} & \multicolumn{1}{c|}{$5.46 \times 10^{-1}$} & $2.13 \times 10^{-7}$ \\
				EfficientNet & \multicolumn{1}{c|}{$2.02$} & \multicolumn{1}{c|}{$2.35$} & \multicolumn{1}{c|}{$1.25$} & $8.48 \times 10^{-7}$\\
				PointNet & \multicolumn{1}{c|}{$2.33$} & \multicolumn{1}{c|}{$0.18$} & \multicolumn{1}{c|}{$2.85$} & $8.62 \times 10^{-7}$\\
				PointNet++ & \multicolumn{1}{c|}{$1.54$} & \multicolumn{1}{c|}{$0.87$} & \multicolumn{1}{c|}{$2.53$} & $6.66 \times 10^{-7}$\\
				\bottomrule
			\end{tabular}
		\end{center}
	\end{small}
\end{table}

\begin{table}[!t]
	\caption{Parameter values for KDE \ie, the values of $h$ ($bw$) for each class and $\lambda$ in the \textit{MAP} formulation.}
	\label{par_map}
	\begin{small}
		\begin{center}
			\begin{tabular}{c|cccc}
				\toprule
				\multirow{2}{*}{\begin{tabular}[c]{@{}c@{}}Neural\\ Network\end{tabular}} & \multicolumn{4}{c}{Parameter} \\ \cline{2-5}
				& \multicolumn{1}{c|}{$h_{Car}$} & \multicolumn{1}{c|}{$h_{Cyc}$} & \multicolumn{1}{c|}{$h_{Ped}$} & $\lambda$ \\ \cline{1-5}
				DenseNet & \multicolumn{1}{c|}{$2.55$} & \multicolumn{1}{c|}{$5.80 \times 10^{-1}$} & \multicolumn{1}{c|}{$1.90$} & $2.58 \times 10^{-7}$ \\
				NasNet & \multicolumn{1}{c|}{$2.48$} & \multicolumn{1}{c|}{$8.81\times 10^{-2}$} & \multicolumn{1}{c|}{$9.96 \times 10^{-2}$} & $3.99 \times 10^{-7}$ \\
				EfficientNet & \multicolumn{1}{c|}{$2.65$} & \multicolumn{1}{c|}{$1.49$} & \multicolumn{1}{c|}{$1.85$} & $1.51\times 10^{-7}$\\
				PointNet & \multicolumn{1}{c|}{$2.51$} & \multicolumn{1}{c|}{$0.55$} & \multicolumn{1}{c|}{$2.51$} & $1.80\times 10^{-7}$\\
				PointNet++ & \multicolumn{1}{c|}{$2.55$} & \multicolumn{1}{c|}{$0.23$} & \multicolumn{1}{c|}{$2.90$} & $1.34\times 10^{-7}$\\
				\bottomrule
			\end{tabular}
		\end{center}
	\end{small}
\end{table}

\begin{table}[!t]
	\caption{Parameter values for the number of bins ($nbins$) of the NHs in the \textit{MAP} formulation.}
	\label{par_bins_map}
	\begin{small}
		\begin{center}
			\begin{tabular}{c|ccc}
				\toprule
				\multirow{2}{*}{\begin{tabular}[c]{@{}c@{}}Neural\\ Network\end{tabular}} & \multicolumn{3}{c}{Parameter} \\ \cline{2-4}
				& \multicolumn{1}{c|}{$nbins_{Car}$} & \multicolumn{1}{c|}{$nbins_{Cyc}$} & \multicolumn{1}{c}{$nbins_{Ped}$} \\ \cline{1-4}
				DenseNet      & \multicolumn{1}{c|}{$13$}  & \multicolumn{1}{c|}{$38$}  & \multicolumn{1}{c}{$25$}  \\
				NasNet        & \multicolumn{1}{c|}{$4$}  & \multicolumn{1}{c|}{$42$} & \multicolumn{1}{c}{$40$} \\
				EfficientNet  & \multicolumn{1}{c|}{$17$} & \multicolumn{1}{c|}{$29$} & \multicolumn{1}{c}{$17$} \\
				PointNet      & \multicolumn{1}{c|}{$7$}  & \multicolumn{1}{c|}{$39$} & \multicolumn{1}{c}{$4$}  \\
				PointNet++ & \multicolumn{1}{c|}{$32$} & \multicolumn{1}{c|}{$25$} & \multicolumn{1}{c}{$5$}  \\
				\bottomrule
			\end{tabular}
		\end{center}
	\end{small}
\end{table}

\section{Concluding Remarks}

The techniques and experimental results described in this paper are based on a proposed probabilistic approach that uses density distributions to model the networks logit values \ie, the top-class scores before the softmax prediction layer. The results reported in this work are very promising, given that \textit{ML} and \textit{MAP} reduced the FPR of the models without the need to retrain the neural networks, while the F-score metric achieved a very small reduction which means the overall classification performance was not compromised.

A potential way to improve the results of the F-score metric by the ML and MAP functions is to adjust the internal parameters of the genetic algorithm (mutation rate, population size, crossover rate, etc.), as well as modifying its cost function.

Finally, a potentially significant aspect that contributed to 
validate the proposed approach for real-world application domains is the use of distinct modalities and sensors, namely by considering RGB images (camera sensor) and $3D$-LiDAR returns point-clouds and range-maps.

\bibliographystyle{IEEEtran}
\bibliography{refs}

\end{document}